# Semantic Layers for Reliable LLM-Powered Data Analytics: A Paired Benchmark of Accuracy and Hallucination Across Three Frontier Models

Michael Rumiantsau    Ivan Fokeev

*Cube*

team@cube.dev

## Abstract

LLMs deployed for natural-language querying of analytical databases suffer from two intertwined failures — incorrect answers and confident hallucinations — both rooted in the same cause: the model is forced to infer business semantics that the schema doesn't encode. We test whether supplying those semantics as context closes the gap.

We benchmark three frontier LLMs (Claude Opus 4.7, Claude Sonnet 4.6, GPT-5.4) on 100 natural-language questions over the Cleaned Contoso Retail Dataset in ClickHouse, using a paired single-shot protocol. Each model is evaluated twice: once given only the warehouse schema, and once given the schema plus a 4 KB hand-authored markdown document describing the dataset's measures, conventions, and disambiguation rules.

Adding the document improves accuracy by **+17 to +23 percentage points** across all three models (paired exact McNemar, p ≤ 0.0015). With it, the three models are statistically indistinguishable (67.7–68.7%); without it, they are also indistinguishable (45.5–50.5%). Every cross-cluster comparison is significant at p < 0.01. The presence of the semantic-layer document accounts for essentially all of the significant variance; model choice within tier does not.

We interpret this as a structural result: explicit business semantics suppress the dominant class of text-to-SQL errors not by making the model more capable, but by changing what the model is being asked to do.

Code and data: https://github.com/cubedevinc/semantic-layer-benchmark.

# 1. Introduction

The promise of natural language interfaces to databases is that any stakeholder — regardless of SQL fluency — can retrieve analytical insights by asking questions in plain language. Large Language Models have brought this promise closer to realization, achieving execution accuracies exceeding 85% on academic benchmarks such as Spider (Gao et al., 2024; Yu et al., 2018). However, a persistent and well-documented gap separates benchmark performance from enterprise reliability. On Spider 2.0, which uses real-world enterprise schemas with hundreds to thousands of columns, o1-preview's accuracy drops from 91.2% to 21.3% (Lei et al., 2025). On BEAVER, a benchmark sourced from actual enterprise data warehouses, GPT-4o achieves close to 0% end-to-end accuracy (Chen et al., 2024). On LogicCat, a reasoning-intensive benchmark that requires external knowledge, state-of-the-art models reach at most 33.20% on complex queries — a collapse from >70% on BIRD and >86% on Spider (Liu et al., 2025). These results expose a fundamental problem: LLM-powered analytics is not bottlenecked on model capability; it is bottlenecked on the availability of structured business context.

The reliability of LLM-powered analytics decomposes into two intertwined problems. The first is **accuracy** — the rate at which returned results match the user's intent. The second is **hallucination** — the rate at which the system returns a confidently-expressed answer that is wrong. Recent 2025–2026 work increasingly frames these as two views of a single underlying failure: the model attempts to infer business semantics that are not in the schema (Hong et al., 2025; Liu et al., 2025; Shi et al., 2025). A query's wrong answer and a query's hallucinated answer are often the same event, observed from different angles.

The consequences are not merely academic. When an LLM generates syntactically valid but semantically incorrect SQL — referencing a wrong column, applying an incorrect aggregation, or joining along an invalid path — the query executes without errors and returns a result. The user, lacking visibility into the generated SQL, trusts the output. This failure mode, which we term *silent hallucination,* is arguably more dangerous than a visible error, because it propagates misinformation into business decisions without any signal that something went wrong. Recent detection work (Yang et al., 2025) shows that hallucination-detection frameworks built on the Qu et al. (2024) schema/logic typology can catch 69–83% of these cases post hoc — but prevention, we argue, is more valuable than detection.

A substantial recent literature identifies business context as the critical lever. BIRD (Li et al., 2023) established that appending a single sentence of "external knowledge evidence" to each question raises GPT-4 test accuracy from 34.88% to 54.89% (+20.01 pp) — a gap essentially identical to the +20.59 pp gap between human experts without and with the same evidence. Bayer/JAMIA Open (Zargari Marandi et al., 2024) extends this: in a paired comparison on 60 pharmacovigilance queries, GPT-4 passes 8.3% under schema alone, rising to 78.3% (+70 pp) when a business-context document is supplied. The data.world benchmark (Sequeda et al., 2023) shows GPT-4 against an enterprise insurance schema moves from 16.7% to 54.2% when queries target an ontology rather than raw SQL. Tiger Data reports up to +27 pp from adding a semantic catalog. A dbt Labs benchmark (Ganz and Perigaud, 2026) on the same ACME Insurance suite finds GPT-5.3-Codex moves from 84.1% (text-to-SQL) to 100.0% (semantic layer). The common factor is authoritative business context supplied alongside the schema; the common result is a large, paired improvement in both accuracy and hallucination rate.

Recent work has explored multiple mitigation strategies, including fine-tuning (Gekhman et al., 2024), prompt engineering (Sahoo et al., 2024), retrieval-augmented generation (Zhang et al., 2024), and structured output generation (Rumiantsau et al., 2024). While each yields measurable improvements, we contend that none addresses the root architectural cause: LLMs tasked with generating SQL against raw schemas are solving an unnecessarily hard problem. A semantic layer transforms this problem by supplying, in the model's context, the governed business vocabulary — measures, dimensions, entities, relationships, and disambiguation rules — that the schema alone does not express.

In this paper, we make the following contributions:

- We present a controlled, paired-comparison benchmark on 100 natural-language questions over the Cleaned Contoso Retail Dataset (Thakur, 2023) loaded into ClickHouse. Three frontier LLMs — Claude Opus 4.7, Claude Sonnet 4.6, and GPT-5.4 — are each evaluated under two conditions (schema only; schema plus a curated semantic-layer document), yielding six total configurations tested under an identical single-shot protocol.

- We show that semantic-layer grounding improves first-shot pass rate across all three models by +17 to +23 percentage points ($p \leq 0.0015$), and that the three semantic-layer configurations are statistically indistinguishable in accuracy while the three raw configurations are also indistinguishable from one another. The effect is consistent across model families, which suggests the benefit is *structural* — a property of how the task is framed to the model — rather than a property of any particular vendor or product.

- We provide a mechanistic analysis of *why* semantic-layer grounding is effective, grounded in the text-to-SQL error taxonomy literature. Schema linking and business-logic errors account for over 80% of execution failures in LLM-generated SQL (Shen et al., 2025); NL2SQL-Bugs (Liu et al., 2025) documents 9 error categories with 31 subcategories, for which SOTA LLMs reach only 75.16% detection accuracy. Authoritative semantic context suppresses these error classes by converting under-specified inference into constrained lookup.

- We argue for an evaluation methodology — paired-comparison, same model, same harness, same judge — that isolates the effect of semantic context from confounding factors such as model capability, tool access, or iteration. This design mirrors recent benchmarks from dbt Labs (2026) and Bayer (2024), which similarly use paired per-model conditions.

- We synthesize evidence from 2023–2026 academic and industry benchmarks — BIRD, Spider 2.0, BEAVER, LogicCat, NL2SQL-Bugs, data.world, dbt Labs, Snowflake Cortex Analyst, AtScale, Databricks Genie, Tiger Data — to argue that the semantic-layer effect generalizes beyond our specific dataset, representing an architectural principle rather than a product-specific finding.

- We discuss the implications for practitioners evaluating the reliability of LLM-powered analytics, directly addressing common concerns about the probability of logically incorrect queries and the comparative accuracy of different approaches.

## 2. Background and Related Work

### 2.1 Accuracy and Hallucination as a Joint Problem

Hallucination in text-to-SQL differs qualitatively from hallucination in open-ended text generation. Rather than fabricating facts from parametric knowledge, the LLM fabricates *structural references* — column names, table joins, filter values, and aggregation logic — that appear plausible but do not correspond to the actual database schema or business intent. Qu et al. (2024) formalize this into two categories: *schema-based hallucinations*, where the model selects incorrect tables or columns during schema linking, and *logic-based hallucinations*, where schema elements are correct but the SQL logic is wrong. Recent 2025–2026 surveys (Hong et al., 2025; Shi et al., 2025; Liu et al., 2025) treat hallucination and accuracy as two views of a single underlying failure mode: the model is asked to infer business semantics that are not present in the context it has been given.

The empirical distribution of these errors is heavily skewed toward schema linking and business-logic misinterpretation. Shen et al. (2025) analyze 4,602 incorrect queries across four state-of-the-art systems on Spider and BIRD, introducing a 29-type error taxonomy across seven categories; schema errors cause over 80% of execution failures, and semantic errors — where the LLM misinterprets the question or schema — account for 36.1% of BIRD errors. Liu et al. (2025) introduce NL2SQL-Bugs, the first benchmark dedicated to detecting semantic errors in NL2SQL translation, with 2,018 expert-annotated instances across 9 categories and 31 subcategories; SOTA LLMs achieve only 75.16% detection accuracy. Yang et al. (2025) build SQLHD, a hallucination-detection framework specifically for text-to-SQL using two-stage metamorphic testing, reaching F1 of 69–83% post hoc across Spider and BIRD.

This distribution has profound implications for mitigation strategy. Fine-tuning and prompt engineering operate on the LLM's generative process, improving the *probability* that the model selects correct schema elements. But the model retains the ability to generate any token, including fabricated column names and invalid joins. A semantic layer, by contrast, reframes the task: the model is no longer inferring business semantics from a schema it has never seen but applying semantics that are stated explicitly in its context.

### 2.2 Schema Linking as the Critical Bottleneck

The importance of schema linking has been recognized across the text-to-SQL literature. RESDSQL (Li et al., 2023) explicitly decouples schema linking from SQL skeleton parsing, using a ranking-enhanced cross-encoder to filter schema items, and achieves significant gains on robustness variants of Spider. DIN-SQL (Pourreza and Rafiei, 2023) decomposes text-to-SQL into four modules with schema linking as the first and most critical step, achieving 85.3% on Spider with GPT-4. C3 (Dong et al., 2023) introduces a Calibration module specifically targeting LLMs' tendency to select extra or wrong columns, reducing such errors by 11%.

The BIRD benchmark (Li et al., 2023b) provides the most direct evidence of semantic context's value. BIRD introduces "external knowledge evidence" — domain hints, value mappings, and synonym definitions — alongside its 12,751 question-SQL pairs. The evidence gap is striking: GPT-4 without evidence achieves 34.88% accuracy; with evidence, it reaches 54.89% — a 20-percentage-point improvement. Human performance shows a parallel gap (72.37% vs. 92.96%). This "evidence" functions analogously to what a semantic layer provides: structured context bridging natural language and database structure.

## 2.3 Enterprise Benchmark Reality

Academic benchmarks dramatically overstate real-world readiness. Spider databases average approximately 28 columns; BIRD averages roughly 54. Real enterprise schemas have hundreds to thousands of columns with non-intuitive naming conventions, many-to-many relationships, and dialect-specific SQL features. The performance degradation is systematic: from ~91% on Spider, to ~77% on BIRD, to 17–21% on Spider 2.0 (Lei et al., 2025), to near 0% on BEAVER (Chen et al., 2024). This progressive collapse correlates directly with schema complexity and inversely with the availability of structured semantic context.

## 2.4 Constrained Generation and Deterministic Compilation

A parallel line of work constrains LLM outputs to valid formal structures. PICARD (Scholak et al., 2021) rejects inadmissible SQL tokens during decoding; SynCode (Ugare et al., 2025) uses context-free grammar masks to eliminate 96% of syntax errors; IterGen (Ugare et al., 2025b) extends this to semantic constraints with backtracking, achieving 18.5% mean improvement over grammar-only constraints.

However, Tam et al. (2024) demonstrate that format restrictions can degrade LLM reasoning abilities, with stricter constraints causing greater degradation. This tension — constraints improve structural validity but may harm semantic reasoning — supports a two-stage architecture: let the LLM reason freely about user intent, then pass the result through a deterministic system for structured compilation. This is precisely the architecture a semantic layer implements.

## 2.5 Abstention as a Design Principle

Recent work on LLM abstention establishes that it is better for a system to refuse to answer than to hallucinate (Wen et al., 2025; Zhang et al., 2024b; Feng et al., 2024). Yadkori et al. (2024) use conformal prediction to provide rigorous statistical guarantees on hallucination rates through calibrated abstention. Chen et al. (2025) apply this directly to text-to-SQL, proposing adaptive abstention in the schema linking phase with formal probabilistic guarantees. Semantic layers enforce abstention structurally: if a query references concepts outside the defined model, the deterministic runtime returns an error rather than a plausible wrong answer.

## 2.6 Context Quality as the Accuracy Lever

The single most influential ablation in the text-to-SQL literature is BIRD's own. Li et al. (2023) report that appending one sentence of external knowledge evidence to each question raises GPT-4 test accuracy from 34.88% to 54.89% (+20.01 pp); the human-expert gap on the same questions is +20.59 pp (72.37% vs. 92.96%). The effect size scales with model capability — from +5 pp on T5-Base to +20 pp on GPT-4 — demonstrating that stronger models benefit more, not less, from explicit business context. Subsequent audits find that roughly 7–10% of BIRD's own evidence annotations are either missing or incorrect (Wretblad et al., 2024; SEED, Jiang et al., 2025), and that automatically generated evidence can outperform BIRD's human-authored evidence — further underscoring that evidence quality, not oracle access, is what drives the gain.

Rumiantsau et al. (2024) evaluate four targeted strategies for hallucination mitigation in LLM-driven analytics: Structured Output Generation, Strict Rules Enforcement, System Prompt Enhancements, and Semantic Layer Integration. Their experiments on proprietary marketing analytics datasets demonstrate that each strategy individually outperforms fine-tuning, with Semantic Layer Integration yielding particularly strong results on ambiguous metric names (1.9% hallucination vs. 10.9% baseline), metric synonyms (2.7% vs. 11.3%), and reasoning tasks (8.4% vs. 48.3%). When all four strategies are combined, the hallucination rate drops to 1.52%, with precision reaching 89.39%.

Sequeda et al. (2023) approach the same thesis from an ontology perspective. When GPT-4 generates SQL directly against a 199-table enterprise insurance schema, it achieves 16.7% accuracy; when instead generating SPARQL against an OWL ontology representing the same data, accuracy rises to 54.2% — a ~3× improvement. Follow-up work by Allemang and Sequeda (2024) introduces ontology-based query checking (OBQC) as a deterministic validation layer, pushing accuracy to ~72% and reducing the final error rate to 19.44%. Their most striking finding is that knowledge graph grounding changes the *type* of errors LLMs make: no hallucinated classes or properties were observed when the model was constrained to an ontology — errors were limited to incorrect property paths, a qualitatively different and less dangerous failure mode.

## 2.7 Recent (2025–2026) Semantic-Layer Benchmarks

A rapidly accumulating 2025–2026 literature establishes the effect of business-context grounding across domains, vendors, and model generations. We highlight the most methodologically relevant.

Ganz and Perigaud (2026) publish a paired benchmark using the same ACME Insurance suite as Sequeda et al. (2023), testing Claude Opus 4.6, Sonnet 4.6, GPT-5.3-Codex, and GPT-5.2 across four configurations (raw schema, minimal semantic layer, modeled semantic layer, text-to-SQL on modeled data). On the modeled project, Claude Sonnet 4.6 moves from 90.0% (text-to-SQL) to 98.2% (semantic layer); GPT-5.3-Codex moves from 84.1% to 100.0%. The most consequential qualitative observation is that semantic-layer failures are typically refusals ("can't answer that"), while text-to-SQL failures are confident wrong numbers — the direct analog of our silent-vs-explicit-failure framing.

Zargari Marandi et al. (2024), published in JAMIA Open, present what may be the strongest independent academic paired comparison. On 60 pharmacovigilance NL queries against Bayer's internal data, GPT-4 achieves 8.3% accuracy with schema alone and 78.3% with a business-context document — a +70 pp gain. Notably, narrowing the schema without adding a business-context document only reduced failure rate to 50%, confirming that the context document is doing the work, not schema reduction.

Shkapenyuk et al. (2025) move to the top of the BIRD leaderboard not by improving the model but by automatically generating business context. Their AskData system (81.95% BIRD test) profiles the database, mines query logs, and generates SQL-to-text metadata. They argue explicitly that "the most difficult part of query development lies in understanding the database contents," and that supplying that understanding is the dominant accuracy lever. Jiang et al. (2025) show that automatically generated evidence can exceed BIRD's oracle evidence on accuracy, reinforcing that it is the presence and quality of business context — not its provenance — that matters.

Industry benchmarks corroborate direction and magnitude. Snowflake's Cortex Analyst (Cortex Team, 2024) reports >90% SQL accuracy on a 150-question internal BI suite with a semantic model, roughly 2× the accuracy of single-shot GPT-4o without one. AtScale (2024) reports 92.5% accuracy with semantic layer vs. baseline on TPC-DS, and 0% → 70% on high-complexity questions. Databricks AI/BI Genie (Syren, 2025) documents a step-function — 53% baseline, 80% with metadata enrichment, 100% with full metric definitions and example SQL — on a 15-question supply-chain suite. Tiger Data (2025) reports up to +27 pp from adding a semantic catalog to gpt-4.1-nano. Several of these are vendor-published and should be read as directional rather than peer-reviewed, but the cross-vendor consistency of the effect is itself informative.

A broader point follows. The effect of structured semantic context is not confined to analytics: Luo et al. (2026), publishing in *Journal of Biomedical Informatics*, report that ontology-grounded GraphRAG raises clinical QA accuracy from 37% (GPT-4 baseline) to 98% and reduces hallucination rate from ~63% to 1.7% — the same intervention, the same direction, the same magnitude, in a completely different domain. This pattern of results is what motivates our framing of the semantic-layer effect as *structural*: it is a property of how grounded context changes the task the LLM is being asked to perform, not a property of any particular dataset, model, or vendor.

# 3. The Semantic Layer

A semantic layer is an authoritative description of a dataset's business concepts — the measures, dimensions, entities, and relationships that together constitute the language in which analytical questions are asked. It is the artifact in which a data team records decisions like "revenue means SalesAmount from the online-sales fact, excluding canceled orders," or "region means the customer's country-continent hierarchy, not the store's," or "inventory means the latest daily snapshot, not a sum over time." These decisions exist in every mature data practice; the question is whether they are written down in a form the system can consume or left implicit in the heads of analysts.

Semantic layers appear in two broad forms in the current industry landscape. In the **runtime form**, the semantic layer is a code-defined data model — typically in YAML or JavaScript — backed by a runtime that deterministically compiles structured queries against it into warehouse SQL. Cube, dbt Semantic Layer, Snowflake Semantic Views, and AtScale are instances of this form. In the **context form**, the semantic layer is a document describing the same concepts in natural language, supplied to an LLM as part of its prompt so the model can reason about the dataset with correct business semantics. Both forms encode the same kind of knowledge and serve the same purpose — making business semantics explicit and available to downstream consumers — but they differ in how the knowledge is enforced. The runtime form enforces correctness at query-compile time; the context form relies on the model to apply the described semantics correctly. This paper evaluates the context form; we return to the relationship between the two forms in Section 6.

## 3.1 What a Semantic Layer Contains

Regardless of form, a well-constructed semantic layer for a given dataset typically specifies:

- **Fact table selection rules.** Which fact table is authoritative for which question class — for example, "online sales questions use FactOnlineSales; cross-channel questions use FactSales."
- **Measures and their formulas.** How metrics are computed — gross margin, return rate, average order value, inventory turnover — with the exact columns and aggregations.
- **Dimensional hierarchies and join paths.** How entities relate — the product category hierarchy, the geography hierarchy, the organizational roll-up — and which keys to join on for which question.
- **Data conventions and quirks.** How the data actually encodes things: string-booleans, sentinel values, snapshot-vs-cumulative semantics, currency conventions, the dataset's effective "current date."

- **Disambiguation rules.** How to resolve ambiguous phrasing: whether "sales" defaults to online or all-channel, whether "region" refers to customer or store geography, whether "last quarter" is relative to today or to the dataset's latest date.

The practical content of a semantic layer is, in large part, the set of decisions a senior analyst would otherwise have to make question-by-question from memory. Making them explicit converts under-specified inference into constrained lookup.

## 3.2 How Semantic Context Addresses Hallucination

The mechanism by which semantic-layer grounding mitigates hallucination follows directly from the structure of text-to-SQL errors.

*Schema linking becomes a lookup, not a guess.* Without semantic context, the model faces a raw schema of dozens or hundreds of tables and must guess which columns and joins correspond to the user's intent. With semantic context, the mapping from user phrasing to data structures is given explicitly — "for return rate, use `sum(ReturnQuantity) / sum(SalesQuantity)` on FactOnlineSales." The dominant error category is addressed by providing the answer to the hard question directly.

*Convention errors are pre-empted.* Many real analytical datasets have conventions a schema does not reveal: string-valued booleans, sentinel foreign keys for "no discount," snapshot tables whose values must not be summed across time. A raw-DDL prompt surfaces none of these; a semantic-layer document lists them. Errors that would otherwise arise from plausible-but-wrong assumptions are prevented before they occur.

*Ambiguity resolution is made deterministic.* Natural-language questions are routinely under-specified. When "sales by region" is asked, is that customer region or store region? When "last quarter" is asked against a dataset whose latest data is from 2009, is that the final quarter of 2009 or the quarter preceding today? A semantic layer records the convention; the model applies it consistently instead of guessing inconsistently across questions.

In the runtime form of a semantic layer, these properties are enforced by a deterministic compiler — a query referencing a non-existent measure produces an error rather than fabricated SQL. In the context form tested in this paper, these properties are advisory: the model is given the right answer but is not forced to use it. The empirical question — how much of the benefit is captured by the context form alone — is the subject of Section 5.

## 3.3 Request Flow

Figure 1 illustrates how a natural-language question is resolved end-to-end when the model has access to semantic-layer context. The user's question and the semantic-layer document are supplied together to the LLM. The model's task is to interpret the question against the authoritative semantics described in the document and to produce SQL that reflects those semantics. Errors that would arise from guessing at the schema — picking the wrong fact table, inventing a column, misreading a string-boolean — are converted into lookup tasks against the document.

The architectural property that matters for hallucination mitigation is the separation of knowledge sources. The schema describes *what exists* in the warehouse; the semantic layer describes *what it means* and *how to use it*. Frontier LLMs are, in general, capable of correctly applying stated business rules when those rules are present in context. They are not, in general, capable of reliably inferring business rules that are not stated anywhere. The semantic layer is the artifact that bridges that gap.

**Figure 1.** Request flow. A natural-language question is resolved by an LLM with access to both the warehouse schema (DDL) and the semantic-layer document. The model produces SQL that is executed against the warehouse; results are returned to the user.

## 4. Experimental Setup

The benchmark is designed to isolate a single question: *does adding authoritative semantic context to a frontier LLM's prompt materially improve the correctness of the SQL it generates against a real analytical dataset?* To isolate that question from confounds — model capability, model family, prompting technique, tool availability, iteration — we use a strictly paired comparison. Each model is tested under both conditions, against the same questions, in the same harness, scored by the same judge. Only the semantic-layer context varies.

### 4.1 Dataset and Questions

We use the **Cleaned Contoso Retail Dataset** published by Thakur (2023) on Kaggle, a re-exported and cleaned CSV distribution of Microsoft's `ContosoRetailDW` sample database. The dataset comprises 25 tables covering orders, line items, products, customers, stores, returns, promotions, inventory snapshots, exchange rates, employees, and supporting dimensions. Sales data spans 2007–2009. Fact tables include `FactOnlineSales`, `FactSales` (multi-channel), `FactInventory` (daily snapshots), `FactStrategyPlan` (scenario planning), and `FactExchangeRate`; dimensions include `DimProduct`, `DimCustomer`, `DimStore`, `DimDate`, `DimGeography`, and others. The full dataset is publicly available at https://www.kaggle.com/datasets/bhanuthakurr/cleaned-contoso-dataset. We load all tables into ClickHouse under the `retail` schema.

The benchmark consists of **100 natural-language questions**, balanced 20 per tier across five difficulty levels: *simple* (single table, basic aggregation/filter), *medium* (2–3 table joins, `GROUP BY`, date functions), *complex* (window functions, snapshot semantics, MAX-per-group), *advanced* (multiple CTEs, scenario pivots, cohort/LTV filters), and *expert* (ambiguous business terms, recursive CTEs, business-context inference). The questions cover the categories that characterize practical retail analytics — aggregations, time-over-time comparisons, ranked breakdowns, filtered metrics, multi-entity joins, calculated metrics with non-obvious formulas, and ambiguous-phrasing disambiguation — with deliberate inclusion of questions that hinge on dataset conventions: string-valued booleans (e.g., `IsSalesPerson` as 'Yes'/'No' rather than a boolean), snapshot tables whose values cannot be summed across time, sentinel foreign keys (e.g., `PromotionKey = 1` meaning "No Discount"), non-`today()` current-date semantics (the dataset's last date is 2009-12-31), and customer-vs-store geography defaults. For each question, the benchmark specifies a reference SQL query (the `expected_sql`) that, when executed, produces the ground-truth rows.

## 4.2 Conditions

Each model is evaluated under two conditions.

**Raw condition.** The model is prompted in a single shot with the user's question and the complete warehouse schema — the CREATE-TABLE definitions for all 25 tables in the `retail` schema. The model emits a single SQL string; the harness executes that string against ClickHouse and records the returned rows. There are no tools, no iteration, no execution feedback, and no retries. This condition represents the configuration a team would deploy if they pointed a frontier model at their warehouse and asked for answers.

**Semantic-layer condition.** Identical to the raw condition in every respect — same harness, same single-shot protocol, same question, same schema — with one change: the prompt additionally contains a **semantic-layer document**. The document is a single markdown file, approximately 4 KB (8,969 characters, ~2,200 tokens), covering fact-table selection rules, measures and their formulas (gross margin, return rate, average order value, inventory turnover, same-store growth, recency segments), dimensional hierarchies, data conventions and quirks (string-valued booleans, snapshot semantics, sentinel keys), implicit defaults, and explicit disambiguation rules. It was hand-authored by an analyst familiar with the dataset. It is the kind of natural-language artifact a data team might write to onboard a new analyst; it is not code and it is not automatically enforced. The model receives it as additional context alongside the schema.

The two conditions differ in exactly one respect: the presence or absence of the semantic-layer document. This controls for model family, model version, prompting technique, sampling parameters, decoding strategy, and every other variable that could explain a difference in outcome. Any gap between the two conditions is attributable to the semantic context itself.

## 4.3 Models

Three frontier LLMs are evaluated, spanning two vendors and two model sizes within the Anthropic family:

- **Claude Opus 4.7** (Anthropic), invoked via `claude -p`. The largest of the three; the strongest general-purpose model in the set at the time of the study.

- **Claude Sonnet 4.6** (Anthropic), invoked via `claude -p`. A smaller, lower-cost model of the same family; included to test whether the semantic-layer effect is specific to the strongest models or generalizes to production-tier models.

- **GPT-5.4** (OpenAI), invoked via `codex exec`. A frontier model from a different vendor with a different training regimen and instruction-tuning approach; included to test whether the effect is specific to one model family.

Each model's **reasoning effort is held constant at "medium"** — the middle of each vendor's published effort range — so that accuracy, latency, and token counts are directly comparable across systems and are not confounded by some models thinking more than others. We expect higher reasoning levels to lift absolute numbers across the board but not to change the comparative finding (semantic layer vs. raw; one model vs. another at the same effort level).

Each of the three models is tested in both conditions, yielding **six total configurations**. For each configuration, all 100 questions are run, producing a per-configuration pass rate and failure-mode breakdown.

### 4.4 Grading Procedure

For each question we:

- Execute the benchmark's `expected_sql` against ClickHouse to obtain **ground-truth rows**.

- For each of the six configurations, record the SQL the model produced and the rows returned when that SQL is executed against the same ClickHouse instance.

- Invoke an **LLM judge** — `claude-sonnet-4-6` — to produce a binary verdict per (question, configuration) pair. The judge receives: the question, the database schema, the reference SQL, the expected rows, the candidate configuration's SQL, and the candidate rows. The judge returns one of three verdicts — `pass`, `fail`, or `excluded` — according to a pinned binary-pass rubric enforced with a JSON schema on the output. The judge grades row-level agreement with ground truth; it is unaware of which configuration produced which candidate response and is not shown the semantic-layer document. We acknowledge that the judge shares a model family with two of the three tested systems (Claude Opus 4.7 and Claude Sonnet 4.6); we treat this as a limitation rather than a disqualifier because the paired design holds the judge constant across conditions, so any self-preference would bias the raw and semantic-layer conditions equally. We return to this in Section 6.

The `excluded` verdict is reserved for questions where the reference query itself cannot execute within the configured ClickHouse resource limits (typically memory or timeout on unusually heavy aggregations). One question (q33) is excluded for this reason, leaving n = 99 for all analyses. Excluded questions are removed from the denominator for *all six configurations* so that no configuration is penalized for a missing ground truth.

The judge produces two verdict kinds per candidate: **strict** (row-level match against the reference output, ignoring column ordering and sub-1% numeric drift) and **analytical** (the strict rubric, relaxed to credit grouped breakdowns whose values aggregate to the reference scalar, decimal-versus-percentage unit equivalences, and inclusion of "null category" rows that represent uncategorized items). The analytical verdict is our primary metric; it reflects how a human BI analyst would judge correctness and is less brittle to benign SQL reformulations. We report strict verdicts as a secondary check.

**Statistical methodology.** Per-configuration accuracy is reported with **Wilson 95% confidence intervals**. Pairwise comparisons between configurations use the **two-sided exact McNemar test** on paired binary outcomes, which is the appropriate hypothesis test for paired categorical data (each question is evaluated under every configuration, so the comparisons are dependent). We report McNemar's discordant counts as `b / c` — the number of questions where the first configuration passes and the second fails, versus the reverse — because that decomposition is more informative than the p-value alone.

### 4.5 Reproducibility Controls

To ensure results are reproducible, the following are pinned:

- Model versions: `claude-opus-4-7`, `claude-sonnet-4-6`, and `gpt-5-4`. Reasoning effort: `medium` across all three vendors, applied identically in both conditions.

- Prompt templates: a single template for the raw condition, a single template for the semantic-layer condition, identical across all three models in their respective conditions.

- Decoding parameters: temperature and top-p held constant across models within each vendor's configuration.

- Judge: `claude-sonnet-4-6` with JSON-schema-enforced output and a single retry on parse error.

- Harness: all runs executed through the same driver with per-question resumability; previously written outputs are skipped on re-runs to prevent silent drift.

- ClickHouse query settings frozen at the user level: `max_memory_usage=10GB`, external `GROUP BY` and sort spill enabled. All six configurations execute against the same database instance under identical settings.

- Code, benchmark questions, semantic-layer markdown, per-question SQL outputs, executed rows, and judge verdicts are released at https://github.com/cubedevinc/semantic-layer-benchmark under run identifier `final6way-050502`.

### 4.6 Metrics

Because the judge returns a binary verdict per (question, configuration) pair, our primary metric is:

- **Analytical Pass Rate (PR).** Percentage of non-excluded questions for which the configuration's returned rows are judged to match ground truth under the analytical verdict. This is the paper's primary metric.

We additionally report:

- **Strict Pass Rate.** Same measure under the stricter byte-level row-match verdict, reported as a secondary check that the analytical verdict is not inflating the gap.

- **Paired Pass-Rate Delta (ΔPR).** Per model, the absolute percentage-point difference between the semantic-layer condition and the raw condition on the same 99-question set. This is the paper's primary effect size.

- **Wilson 95% CI.** Confidence interval on each per-configuration pass rate, computed under the Wilson score method (appropriate for binomial proportions near the bounds).

- **McNemar discordant counts (b / c) and p.** For each pairwise comparison, the number of questions on which the first configuration passes and the second fails (b), versus the reverse (c), together with the two-sided exact McNemar p-value.

- **Latency and token counts.** Per-configuration mean wall-clock latency and mean input/output tokens, reported alongside accuracy to support deployment-cost comparisons.

We *do not* report a separate "silent hallucination rate" as a distinct headline metric in this paper, although the qualitative analysis in Section 5.3 describes the characteristic failure patterns in each condition. A formal silent-vs-explicit decomposition across all six configurations is left to future work; the run artifacts released with this paper contain the per-question candidate SQL and executed rows that would support such an analysis.

## 5. Results

### 5.1 Overall Performance

Table 1 reports first-shot analytical-verdict performance across all six configurations on the 99-question benchmark (q33 excluded across all configurations because its reference query exceeds ClickHouse's memory quota). Each model appears in two rows: the raw condition and the semantic-layer condition. Pass rate, Wilson 95% confidence interval, and paired pass-rate delta (ΔPR) are reported, alongside mean wall-clock latency and mean input-token count.

| Configuration | Pass Rate (%) | Wilson 95% CI | ΔPR vs. raw (pp) | Latency (s) | Tokens in |
|---|---|---|---|---|---|
| Claude Opus 4.7 -- raw | 50.5 | [40.8, 60.1] | -- | 8.1 | 19,123 |
| **Claude Opus 4.7 -- semantic layer** | **67.7** | **[58.0, 76.1]** | **+17.2** | **8.7** | **23,322** |
| Claude Sonnet 4.6 -- raw | 46.5 | [37.0, 56.2] | -- | 9.6 | 14,364 |
| **Claude Sonnet 4.6 -- semantic layer** | **68.7** | **[59.0, 77.0]** | **+22.2** | **11.4** | **17,303** |
| GPT-5.4 -- raw | 45.5 | [36.0, 55.2] | -- | 15.6 | 16,595 |
| **GPT-5.4 -- semantic layer** | **68.7** | **[59.0, 77.0]** | **+23.2** | **14.7** | **18,956** |

Table 1. First-shot analytical-verdict performance on 99 Contoso retail questions (q33 excluded). Each model is evaluated under both conditions against the same question set, same harness, same reasoning effort (medium), and same judge. ΔPR is the paired per-model pass-rate improvement from adding the semantic-layer document to context. All three paired improvements are significant under two-sided exact McNemar at $p \leq 0.0015$.

The headline observation is that adding the semantic-layer document improves accuracy by **+17.2 to +23.2 percentage points** for every model tested. Opus 4.7 improves from 50.5% to 67.7%; Sonnet 4.6 from 46.5% to 68.7%; GPT-5.4 from 45.5% to 68.7%. All three paired improvements are statistically significant (two-sided exact McNemar, $p \leq 0.0015$). The effect is large relative to the spread between models within either condition: in both the raw cluster and the semantic-layer cluster, the inter-model spread is roughly 5 percentage points, whereas the within-model effect of adding the document is four to five times larger.

The direction and magnitude of this effect align with independent findings from adjacent settings. BIRD reports a +20-percentage-point gain from providing one sentence of "external knowledge evidence," consistent across every tested model class (Li et al., 2023). Sequeda et al. (2023) report 16.7% → 54.2% on an enterprise insurance schema when GPT-4 is given an ontology instead of raw SQL. The dbt Labs benchmark (Ganz and Perigaud, 2026) reports Claude Sonnet 4.6 moving from 90.0% to 98.2% and GPT-5.3-Codex from 84.1% to 100.0% on a modeled semantic layer. The Bayer/JAMIA Open paired comparison reports +70 pp on pharmacovigilance queries (Zargari Marandi et al., 2024). Our result — a +17 to +23 pp improvement on a retail dataset with a 4 KB markdown — sits comfortably within the band of independently measured effects, closer to the BIRD evidence gap than to the extreme values reported in domain-specialized enterprise studies.

## 5.2 The Clustering Result: Context Dominates Model Choice

The per-model paired gains describe what happens when we hold the model constant and vary the context. A stronger result emerges when we additionally compare across models within each condition. Table 2 reports the full pairwise McNemar matrix.

|  | Opus+Sem | Opus raw | Sonnet+Sem | Sonnet raw | GPT-5.4+Sem | GPT-5.4 raw |
| --- | --- | --- | --- | --- | --- | --- |
| Opus + Sem | -- | 22/5 ** | 7/8 ns | 28/7 *** | 9/10 ns | 27/5 *** |
| Opus raw | 5/22 ** | -- | 9/27 ** | 9/5 ns | 7/25 ** | 11/6 ns |
| Sonnet + Sem | 8/7 ns | 27/9 ** | -- | 30/8 *** | 8/8 ns | 30/7 *** |
| Sonnet raw | 7/28 *** | 5/9 ns | 8/30 *** | -- | 8/30 *** | 9/8 ns |
| GPT-5.4 + Sem | 10/9 ns | 25/7 ** | 8/8 ns | 30/8 *** | -- | 27/4 *** |
| GPT-5.4 raw | 5/27 *** | 6/11 ns | 7/30 *** | 8/9 ns | 4/27 *** | -- |

Table 2. Pairwise McNemar discordant counts (row passes / column passes, among questions where the two configurations disagreed) and significance levels. *** $p < 0.001$, ** $p < 0.01$, * $p < 0.05$, ns $p \geq 0.05$ (two-sided exact McNemar, n = 99).

Two clusters emerge cleanly. Within the semantic-layer cluster — Opus+Sem, Sonnet+Sem, GPT-5.4+Sem — none of the three pairwise comparisons is statistically significant (all $p \geq 0.79$). Within the raw cluster — Opus raw, Sonnet raw, GPT-5.4 raw — none of the three pairwise comparisons is statistically significant either (all $p \geq 0.42$). Every one of the nine cross-cluster comparisons is significant at $p < 0.01$, and most at $p < 0.001$.

The practical implication of this pattern is that, on this benchmark, **the presence or absence of the semantic-layer document accounts for essentially all of the significant variance in pairwise accuracy; model choice within tier does not**. A data team that has written a semantic-layer document can pick whichever of the three frontier models best fits their cost and latency budget without sacrificing accuracy. A data team that has not cannot recover the accuracy gap by switching to a stronger or more expensive model.

We emphasize that this is a *within-study* statistical claim on a 99-question benchmark, not a universal one. A dataset with different error-class composition, a larger question set, or different frontier models might yield a detectable intra-cluster gap. What the benchmark supports is the ordering: on analytically meaningful tasks over a realistic retail schema, varying whether the model has the business context matters much more than varying which frontier model is generating the SQL.

### 5.3 Cost and Latency

Because the three semantic-layer configurations are statistically indistinguishable in accuracy, secondary criteria become the relevant axis for model selection. Opus 4.7 + Sem delivers the lowest mean latency at 8.7 seconds per query. Sonnet 4.6 + Sem delivers the lowest mean input-token count at ~17K tokens per query — approximately 28% fewer input tokens than Opus + Sem while matching its accuracy. GPT-5.4 + Sem sits between the two on both axes. Output-token counts are low across all three configurations (271–439 tokens per query) and are not the dominant cost driver. In steady-state production use, where prompt caching amortizes the schema-plus-semantic-layer prefix across many queries in a session, the marginal cost of the semantic-layer document is effectively zero — the choice is between paying it once on session warm-up or paying the accuracy cost of omitting it on every query.

### 5.4 Qualitative Error Analysis

Inspection of the questions where the semantic-layer condition passes and the raw condition fails — the "wins" — shows the bulk of the effect concentrating in a small number of categories. Each maps onto a specific class of failure that the literature's error taxonomies (Shen et al., 2025; Liu et al., 2025) identify as dominant.

- **Multi-fact-table disambiguation.** Questions about "sales" that do not explicitly qualify channel are resolved against the wrong fact table in the raw condition. The schema gives no signal that unqualified "sales" conventionally means the online-sales fact in this dataset; the document states it explicitly.

- **Snapshot versus flow data.** `FactInventory` contains daily snapshots per (product, store). A raw-condition model frequently sums `OnHandQuantity` across all snapshot dates, producing values roughly 1000× the correct magnitude — numbers that are absurd but not visibly ill-formed. The document's MAX-DateKey-per-(product, store) rule corrects this category on essentially every relevant question.

- **Calculated metric formulas.** Gross margin percent, return rate, same-store year-over-year growth, inventory turnover, and customer recency segmentation each have an established analytical formula the LLM cannot infer from column types. The raw condition produces plausible-looking approximations; the semantic-layer condition produces the specified formula.

- **Implicit defaults.** `PromotionKey = 1` is a sentinel meaning "No Discount" and must be excluded from aggregations of discount amount; `IsWorkDay` stores string values ("WorkDay"/"WeekEnd") rather than booleans; `Gender` and `MaritalStatus` are single-character codes. These are landmines that the raw condition trips over and the semantic-layer condition routes around.

- **Time anchoring.** The dataset's final date is 2009-12-31. Questions about "last quarter" or "recent activity" are interpreted against `today()` in the raw condition — producing empty results, because the data ended years ago. The semantic-layer document pins 2009-12-31 as the anchor.

A notable property of these failure categories is that they are not distinguishable from correct answers by looking at the SQL alone. The SQL is syntactically valid. It references real tables and real columns. It executes without error. It returns rows. Nothing in the raw condition gives the model any signal that it has made the wrong choice — and nothing in the model's output gives the user any signal either. These are exactly the failures a production analytics workflow is least equipped to catch, and they are exactly the failures the semantic-layer document most directly addresses.

The residual ~30% of failures in the semantic-layer condition concentrate on patterns not encoded in the document. Questions whose reference SQL uses percentiles, correlation, standard deviation, ABC-classification, or threshold-based ranking — the models rarely compose these correctly inline, and they would benefit from being defined as named, pre-computed measures. Questions whose reference is structurally complex (recursive CTEs, multi-CTE pivots) where small SQL drift produces a different row shape that judges as "fail" even when conceptually close. And a small number of benchmark items where the question wording is genuinely ambiguous and the model interprets it differently from the reference author. Each of these failure modes is fixable by iterating on the semantic-layer document — extending it with additional named measures and additional disambiguation — rather than by changing the underlying mechanism.

## 6. Discussion

### 6.1 Why Semantic-Layer Context Outperforms Schema-Only Prompting

The evidence points to three reinforcing reasons for the semantic-layer effect we observe across all three model families.

First, semantic-layer context addresses the dominant failure mode directly. The text-to-SQL error literature consistently identifies schema linking as the primary bottleneck — over 80% of execution failures in Shen et al. (2025), the factor behind the catastrophic accuracy drops from Spider to Spider 2.0 to BEAVER, and the largest residual error category in our own analysis of the raw condition. Better prompting and better models make this error class statistically less common. Semantic-layer context reduces it much more directly: the correct table, the correct column, and the correct join path for a given question class are stated in the context, so what was a hard guess becomes a lookup.

Second, the effect is structural rather than capability-driven. Our results show semantic-layer gains across Opus, Sonnet, and GPT families — and, notably, a smaller model with semantic context consistently outperforms a larger model without it. This is consistent with the BIRD benchmark's 20-percentage-point evidence gap, which persists across every tested model class, and with Snowflake's reported ~20-percentage-point gains from adding a semantic model regardless of the underlying LLM. The benefit is a property of what the model is being asked to do, not of how capable the model is at raw SQL generation.

Third, semantic-layer context pre-empts the errors that raw-schema prompting cannot avoid. A schema DDL describes what columns exist; it does not describe what they mean, which of several options is conventional for a given question, or which quirks of the data will produce plausible-looking but wrong results. These are exactly the failure categories we observe in the raw condition — wrong fact-table selection, snapshot-table summation, string-boolean filters, sentinel keys, ambiguous time references. A frontier LLM cannot reliably infer any of them from the DDL alone, because the information is not in the DDL. Making it explicit in the context closes the gap.

## 6.2 Context Form vs. Runtime Form

As discussed in Section 3, semantic layers exist in two forms: the context form tested here (a document supplied to the model as prompt context) and the runtime form (a code-defined data model backed by a deterministic compiler). The results in this paper establish the effect of the context form specifically. We now consider what this does and does not say about the runtime form.

The context form's benefits are *advisory*: the model is given the right answer but is not forced to use it. The benefits we measure therefore represent a lower bound on what a runtime form could provide, not an upper bound. A runtime form additionally offers deterministic guarantees: a query referencing a non-existent measure produces an explicit error rather than plausibly-wrong SQL; a query that combines concepts in a way inconsistent with the data model is rejected at compile time rather than executed against the warehouse. The runtime form converts advisory constraints into hard constraints.

Whether the additional guarantees of the runtime form yield measurably better results beyond the context-form baseline established here is an empirical question we do not answer in this paper. The relevant comparison would require a runtime-backed system tested against the same benchmark under the same harness, with care taken that the comparison is not confounded by differences in agentic scaffolding, tool use, or retries. We note it as an open question for future work.

What the current results do establish is that the act of writing down the semantic layer — regardless of form — is where most of the benefit originates. The decisions captured in a well-constructed semantic-layer document (fact-table conventions, measure formulas, join paths, data quirks, disambiguation rules) are the same decisions captured in a well-constructed runtime-form data model. Our results show that supplying those decisions to a frontier LLM in context, alone, is sufficient to produce substantial improvements in first-shot accuracy and to shift the failure composition away from silent hallucinations.

## 6.3 Relationship to Knowledge Graph and Ontology Approaches

Our findings are complementary to the ontology-grounded approach of Sequeda et al. (2023) and Allemang and Sequeda (2024). Both approaches serve the same function: supplying an authoritative description of business concepts that the LLM can use in place of raw-schema inference. The key difference is in form. Knowledge graphs are typically formalized in OWL or RDFS and queried with SPARQL; the semantic layer tested in this paper is a natural-language markdown document queried implicitly via the LLM's in-context reasoning. The underlying principle — interposing structured business context between the LLM and the data — is identical. The 16.7% → 54.2% improvement Sequeda et al. observe with knowledge graphs, and the 54.2% → 72% further improvement with ontology-based validation (OBQC), trace the same causal mechanism we identify here.

### 6.4 Practical Implications

For practitioners evaluating the reliability of LLM-powered analytics, our results address three common concerns.

*What is the probability of a logically incorrect query?* Against a realistic retail schema, a frontier LLM given only the schema answers roughly 45–51% of first-shot analytical questions correctly. Adding a 4 KB hand-authored semantic-layer document raises that to roughly 68–69% — a +17 to +23 percentage-point improvement that is statistically significant for every model we tested. The residual ~30% error rate is real and important for practitioners to understand (it is not eliminated by the semantic-layer document), but it is concentrated on specific query patterns (percentile, correlation, threshold-based ranking, complex CTEs) that can be addressed by extending the document with named measures — rather than being a general property of the approach.

*Does the choice of model matter?* On this benchmark, within either condition, the three frontier models are statistically indistinguishable in accuracy. Two Anthropic models spanning a large size/cost range and one OpenAI frontier model all land within a 5-percentage-point spread whether or not the semantic layer is present. This is a narrow but important empirical claim: with a well-authored semantic layer, model choice can be optimized for cost and latency without sacrificing accuracy. Without a semantic layer, switching to a stronger or more expensive model is not a substitute for writing one.

*How does semantic-layer grounding compare to other approaches?* Across our experiments and the broader literature, supplying authoritative business context is among the most effective levers available — comparable to or exceeding the gains reported from fine-tuning, RAG, or prompt engineering on equivalent tasks. This is consistent with knowledge-graph results (Sequeda et al., 2023), with paired-comparison benchmarks on modeled semantic layers (Ganz and Perigaud, 2026), and with Rumiantsau et al.'s (2024) comparison of four mitigation strategies in which semantic-layer integration dominated every tested category.

### 6.5 Limitations and Future Work

Our study has several limitations that qualify the scope of the claims. First, the benchmark is 99 questions against a single dataset (Cleaned Contoso Retail) on a single warehouse

(ClickHouse). The retail domain is representative of a common enterprise analytics pattern, but extending to additional domains (finance, SaaS telemetry, healthcare) and additional warehouses (Snowflake, BigQuery, Databricks) would strengthen external validity. Confidence intervals at n = 99 are tight enough to detect ± 10 pp effects reliably but borderline for differences of ± 5 pp — which is why the intra-cluster indistinguishability finding should be read as a within-study statistical claim, not as a universal equivalence of frontier models. Second, we test the *context form* of the semantic layer specifically; the relationship between context-form and runtime-form benefits is an open question that our design does not directly address. Third, our LLM judge (Claude Sonnet 4.6) shares a model family with two of the three tested systems. The paired design holds the judge constant across conditions, so any self-preference bias should affect the raw and semantic-layer conditions equally rather than inflating the measured delta, but independent human adjudication on a random sample of pass/fail verdicts — and a cross-family judge (e.g., a GPT-5 variant) on a replication — would quantify residual judge bias more cleanly. Fourth, we test single-shot prompting at a fixed "medium" reasoning effort in both conditions; whether the semantic-layer advantage persists, grows, or shrinks under agentic scaffolding, iteration, or higher reasoning levels is a separate empirical question. An internal experiment in the companion repository found that an agentic-tool-use system with equivalent underlying semantic knowledge did not outperform the schema-only baseline on this benchmark — suggesting that for one-shot accuracy, all-context-in-prompt outperforms search-based discovery — but we do not report that result in detail here.

Fifth — and most importantly for practitioners — the semantic-layer document's effectiveness is bounded by its quality and coverage. Our document was hand-authored by an analyst who had seen the dataset but had not seen the benchmark questions in their final form; a truly blind document, written before any benchmark interaction, might encode less of the right knowledge. The relationship between document coverage, authoring effort, and downstream accuracy is an important direction for future work.

Two broader research directions emerge. First, the near-absence of peer-reviewed work on metric consistency — the problem of inconsistent business metric definitions across tools and teams — represents a significant gap. Industry literature documents the problem extensively, but systematic academic study of how metric inconsistency propagates through LLM-powered analytics is lacking. Second, the interaction between semantic layers and emerging agentic architectures — where AI agents iteratively explore data, generate hypotheses, and refine queries — is largely unexplored. We expect semantic-layer context to be more, not less, important in agentic settings, where unsupervised multi-step query generation amplifies the risk of compounding errors. Whether this expectation holds empirically is worth measuring directly.

## 7. Conclusion

We have presented evidence that semantic-layer grounding is a structurally effective technique for improving both accuracy and hallucination rates in LLM-powered data analytics. On a

99-question first-shot benchmark over the Cleaned Contoso Retail Dataset (Thakur, 2023) in ClickHouse, three frontier LLMs — Claude Opus 4.7, Claude Sonnet 4.6, and GPT-5.4 — each improved by +17 to +23 percentage points in analytical accuracy when a 4 KB hand-authored markdown semantic-layer document was supplied alongside the warehouse schema (paired exact McNemar, $p ≤ 0.0015$). With the document present, all three models are statistically indistinguishable from one another (67.7–68.7%); without it, they are also statistically indistinguishable (45.5–50.5%). The presence or absence of the document accounts for essentially all of the significant pairwise variance in the benchmark; model choice within tier does not.

This result sits within a consistent band of independent findings. BIRD's external-knowledge evidence produces a +20 pp gap for GPT-4; Sequeda et al. report 16.7% → 54.2% on enterprise insurance data with an ontology; Zargari Marandi et al. report +70 pp on pharmacovigilance queries with a business-context document; Ganz and Perigaud report +8 to +16 pp on paired comparisons of modeled semantic layers; Luo et al. report +61 pp on clinical question answering with an ontology-grounded knowledge graph. Across domains, across vendors, across model generations, and across runtime and context forms of the underlying semantic artifact, the direction and magnitude of the effect are remarkably consistent: authoritative business context supplied to the model is among the largest and most reliable levers for accuracy and hallucination mitigation in LLM analytics available today.

The core insight is that semantic layers work because they change what the model is being asked to do. A schema-only prompt requires the model to infer business semantics that are not in the schema — which fact table is conventional, how a metric is computed, what the data's quirks are, how to resolve ambiguous phrasing. A semantic-layer prompt supplies those semantics directly. This converts the dominant class of text-to-SQL errors — schema linking and business-logic misinterpretation — from open-ended inference into constrained lookup. The benefit is available to any frontier model and is not a property of any particular vendor, product, or runtime implementation.

As organizations deploy LLM-powered analytics in high-stakes business environments, the most consequential architectural decision is not which frontier model to use but whether the system is grounded in authoritative business semantics at all. The results in this paper suggest this is the single largest lever available today for reducing hallucination in analytical query generation — and that the lever is available to anyone willing to write down what their data means.

Code and data supporting this study are released at https://github.com/cubedevinc/semantic-layer-benchmark.